\pdfoutput=1

\documentclass[11pt]{article}

\usepackage{ACL2023}

\usepackage{times}
\usepackage{latexsym}
\usepackage{soul}
\usepackage[T1]{fontenc}

\usepackage[utf8]{inputenc}

\usepackage{microtype}

\usepackage{inconsolata}

\usepackage[normalem]{ulem}
\usepackage{subcaption}
\usepackage{graphicx}
\usepackage{tabularx}
\usepackage{breakcites}
\usepackage{booktabs}
\usepackage{multirow}
\usepackage{amsmath,amsthm,mathtools}
\usepackage{hyperref}

\title{Causal Intervention for Abstractive Related Work Generation}

\author{Jiachang Liu \\ 
    Beijing Institute of Technology \\ 
    Beijing, China \\ 
    \texttt{jc\_liu@bit.edu.cn} \\ \And
    Qi Zhang \\ 
    Tongji University \\ 
    Shanghai, China \\ 
    \texttt{zhangqi\_cs@tongji.edu.cn} \\\And
    Chongyang Shi \\ 
    Beijing Institute of Technology \\ 
    Beijing, China \\
    \texttt{cy\_shi@bit.edu.cn} 
    \AND
    Usman Naseem \\ 
    University of Sydney \\ 
    Sydney, Australia \\\And
    Shoujin Wang \\ 
    University of Technology Sydney \\ 
    Sydney, Australia \\\And
    Ivor Tsang \\ 
    A*STAR \\ 
    Singapore 
    }

\begin{document}
\maketitle
\begin{abstract}
Abstractive related work generation has attracted increasing attention in generating coherent related work that better helps readers grasp the background in the current research. 
However, most existing abstractive models ignore the inherent causality of related work generation, leading to low quality of generated related work and spurious correlations that affect the models' generalizability. 
In this study, we argue that causal intervention can address these limitations and improve the quality and coherence of the generated related works. 
To this end, we propose a novel {\itshape \textbf{Ca}usal Intervention \textbf{M}odule for Related Work Generation} (CaM) to effectively capture causalities in the generation process and improve the quality and coherence of the generated related works.
Specifically, we first model the relations among 
sentence order, document relation, and transitional content 
in related work generation using a causal graph. 
Then, to implement the causal intervention and mitigate the negative impact of spurious correlations, we use \textit{do}-calculus to derive ordinary conditional probabilities and identify causal effects through CaM. 
Finally, we subtly fuse CaM with Transformer to obtain an end-to-end generation model. 
Extensive experiments on two real-world datasets show that causal interventions in CaM can effectively promote the model to learn causal relations and produce related work of higher quality and coherence.
\end{abstract}

\section{Introduction}
A comprehensive related work necessarily covers abundant reference papers, which costs authors plenty of time in reading and summarization and even forces authors to pursue ever-updating advanced work~\cite{hu-wan-2014-automatic}. Fortunately, the task of related work generation emerged and attracted increasing attention from the community of text summarization and content analysis in recent years~\cite{chen-etal-2021-capturing, target-aware-2022}. 
Related work generation can be considered as a variant of the multi-document summarization task~\cite{li2022automatic-meta-review}. 
Distinct from multi-document summarization, related work generation entails comparison after the summarization of a set of references and needs to sort out the similarities and differences between these references~\cite{agarwal-etal-2011-scisumm}.

Recently, various abstractive text generation methods have been proposed to generate related work based on the abstracts of references. For example, \newcite{xing-etal-2020-automatic} used the context of citation and the abstract of each cited paper as the input to generate related work. \newcite{ge-etal-2021-baco} encoded the citation network and used it as external knowledge to generate related work. \newcite{target-aware-2022} proposed a target-aware related work generator that captures the relations between reference papers and the target paper through a target-centered attention mechanism. Equipped with well-designed encoding strategies, external knowledge, or novel training techniques, these studies have made promising progress in generating coherent related works.

However, those models are inclined to explore and exploit spurious correlations such as high-frequency word/phrase patterns, writing habits, or presentation skills, building superficial shortcuts between reference papers and the related work of the target paper. 
Such spurious correlations may affect or even harm the quality of the generated related work, especially under the distribution shift between the testing set and training set. 
This is because spurious correlations different from genuine causal relations may not intrinsically contribute to the related work generation and easily cause the robustness problem and impair the models' generalizability~\cite{invariant-spurious-corr}.

\begin{figure}
	\centering
	\includegraphics[width=1\linewidth]{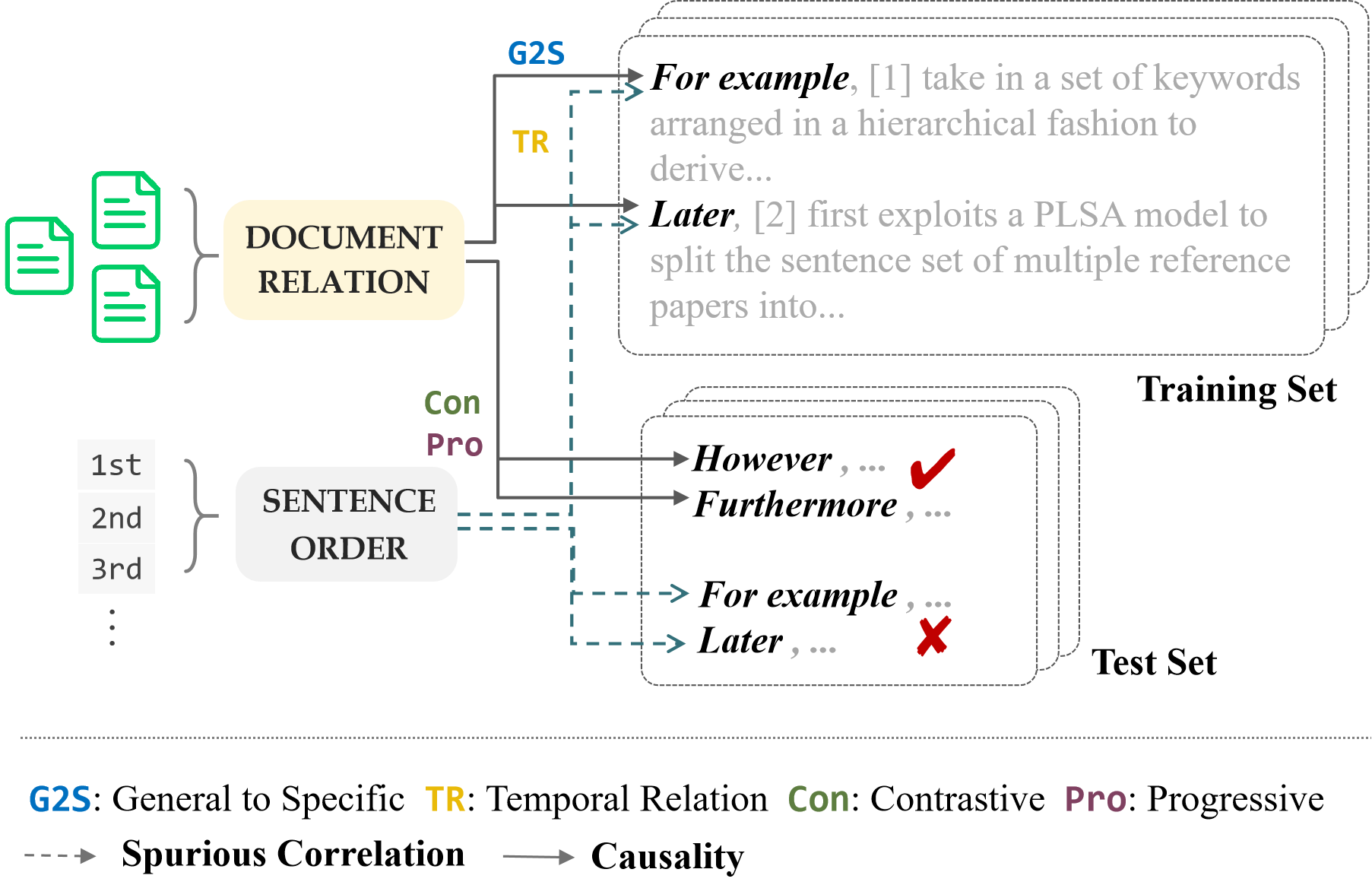}
	\caption{An illustration of the effect difference between causality (solid arrows) and spurious correlations (dashed arrows) in related work generation.
    }
	\label{fig:problem}
	\vspace{-3mm}
\end{figure}

Figure~\ref{fig:problem} illustrates the difference between causality and spurious correlation. The phrases "for example" and "later" are often used to bridge two sentences in related work. Their usage may be attributed to writers' presentation habits about organizing sentence orders or the reference document relations corresponding to the sentences. 
Ideally, a related work generation model is expected to learn the reference relation and distinguish it from the writing habits. However, the generation model easily captures the superficial habitual sentence organization (spurious correlation) instead of learning complex semantic reference relations (causality), especially when the habitual patterns frequently occur in the training set. In this case, the transitional phrases generated mainly based on writing habits are likely to be unsuitable and subsequently affect the content generation of related work during testing when the training and testing sets are not distributed uniformly. 

Fortunately, causal intervention can effectively remove spurious correlations and focus on causal correlations by intervening in the learning process. It not only observes the impact of the sentence order and document relation on generating transitional content but probes the impact of each possible order on the whole generation of related work, thereby removing the spurious correlations~\cite{pearl-intervention}.
Accordingly, causal intervention serving as an effective solution allows causal relations to exert a greater impact and instruct the model to produce the correct content.

\begin{figure}[!t]
	\centering
	\includegraphics[width=1\linewidth]{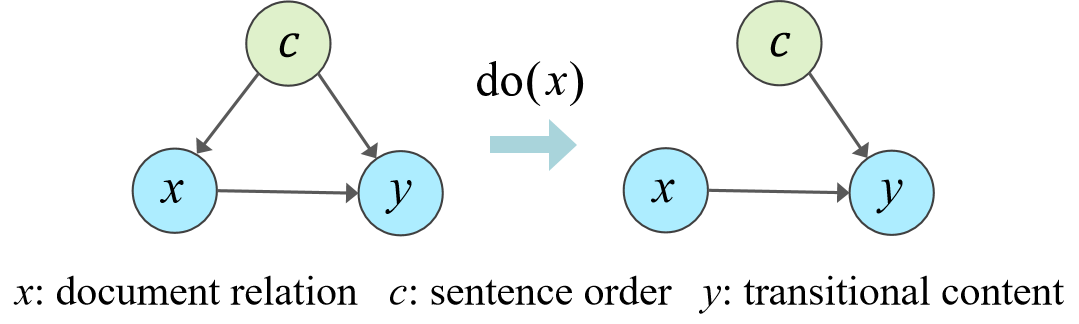}
	\caption{Causal graph $G$ for related work generation. By applying $do$-calculus, path $c \to x$ is cut off and the impact of spurious correlation $c \to x \to y$ is mitigated.}
	\label{fig:causal graph}
	\vspace{-3mm}
\end{figure}

To address the aforementioned gaps in existing work for related work generation, we propose a 
{\textbf{Ca}usal Intervention \textbf{M}odule for Related Work Generation} (CaM). CaM can effectively remove spurious correlations by performing the causal intervention, therefore producing related work with high quality. 
Specifically, we first model the relations among sentence order, document relation, and transitional content in related work generation and figure out the confounder that raises spurious correlations (see Figure \ref{fig:causal graph}). Then, we implement causal intervention via the proposed CaM that consists of three components:
1) \textit{Primitive Intervention} cuts off the connection that induces spurious correlations in the causal graph 
by leveraging \textit{do}-calculus and \textit{backdoor criterion} \cite{pearl-intervention}; 
2) \textit{Context-aware Remapping} smoothens the distribution of intervened embeddings and injects contextual information; and
3)\textit{ Optimal Intensity Learning} learns the best intensity of overall intervention by controlling the output from different parts. 
Finally, we strategically fuse CaM with Transformer \cite{attentionallyouneed} to deliver an end-to-end causal related work generation model. Our main contributions are as follows:\vspace{-2mm}

\begin{itemize}
\item To the best of our knowledge, this work is the first attempt to introduce causality theory into related work generation task.\vspace{-3mm}
\item We propose a novel {\textbf{Ca}usal Intervention \textbf{M}odule for Related Work Generation} (CaM) which implements causal intervention to mitigate the impact of spurious correlations. CaM is subtly fused with {Transformer} to derive an end-to-end causal model, enabling the propagation of intervened information.\vspace{-3mm}
\item Extensive experiments on two related work generation datasets demonstrate that our model outperforms the state-of-the-art approaches and verifies the effectiveness and rationality of bringing causality theory into the related work generation task.
\end{itemize}

\section{Problem Formulation}
Given a set of reference papers $D = \{{r}_1, ..., {r}_{|D|}\}$, we assume the ground truth related work $Y = (w_1, w_2,..., w_M)$, 
where ${r}_i = (w^i_1, w^i_2,..., w^i_{|r_i|})$ 
denotes a single cited paper, $w^i_j$ is the $j$-th word in ${r}_i$, and $w_j$ is the $j$-th word in related work ${Y}$. Generally, the related work generation task can be formulated as generating a related work section $\hat{{Y}} = (\hat{w}_1, \hat{w}_2,..., \hat{w}_{\hat{M}})$ based on the reference input $D$ and minimizing the difference between ${Y}$ and $\hat{{Y}}$. Considering that the abstract section is usually well-drafted to provide a concise paper summarization~\cite{hu-wan-2014-automatic}, we use the abstract section to represent each reference paper.

\begin{figure*}[!t]
	\centering
	\includegraphics[width=0.98\linewidth]{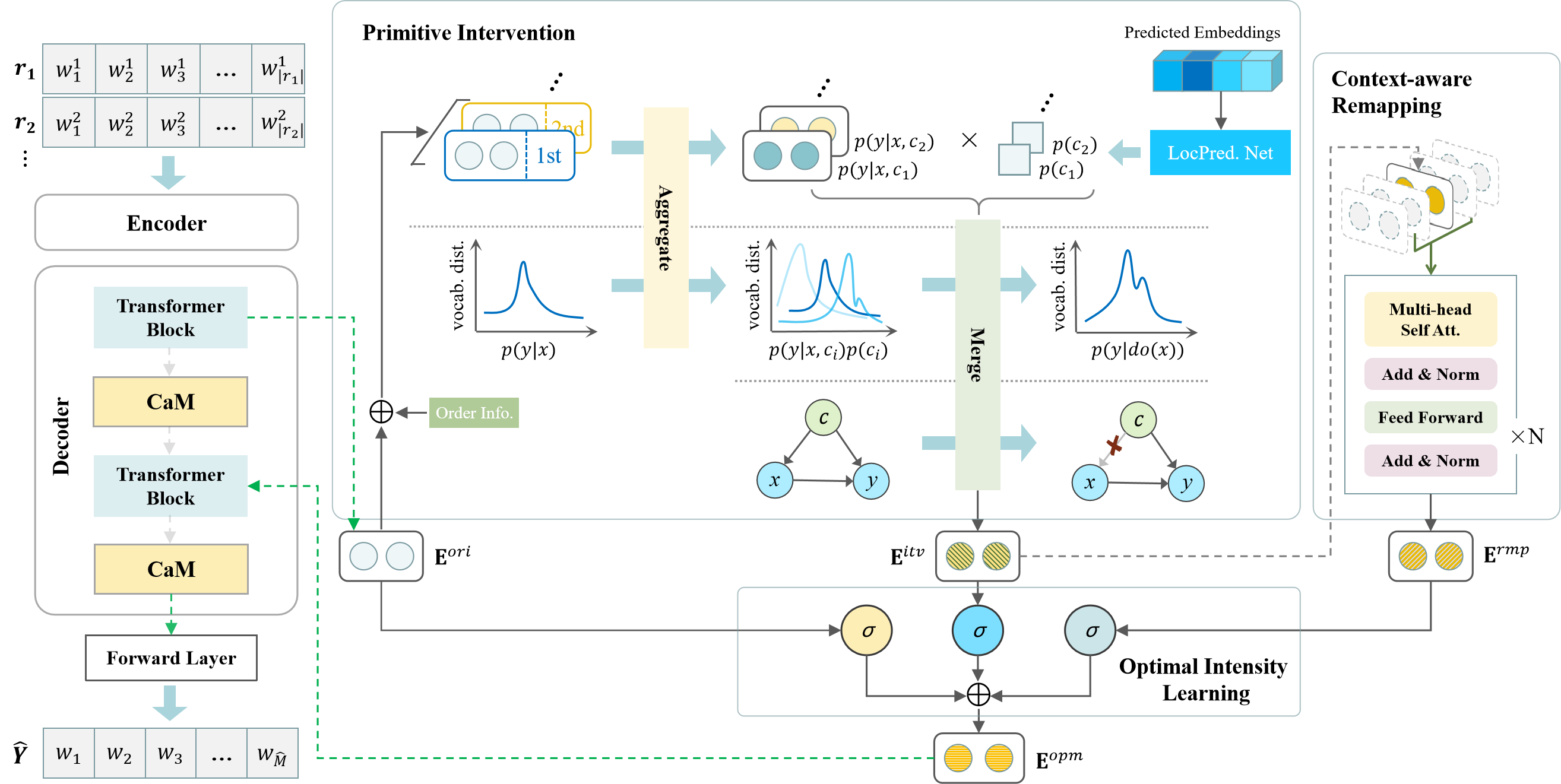}
	\caption{The structure of CaM fused with the Transformer in the decoder. CaM consists of three parts: Primitive Intervention, Context-aware Remapping and Optimal Intensity Learning.}
	\label{fig:overall model}
	\vspace{-2mm}
\end{figure*}

\section{Methodology}

We first analyze the causalities in related work generation, identify the confounder that raises spurious correlations, and use the causal graph to model these relations. Then, we introduce how CaM is designed to enhance the quality of related work through causal intervention. Finally, we describe how CaM, as an intervention module, is integrated with the Transformer to influence the entire generation process. The overall structure of our model is shown in Figure \ref{fig:overall model}.


\subsection{Causal Modeling for Related Work Generation}
We believe that three aspects play significant roles in related work generation for better depicting the relations between different references, namely, sentence order $c$, document relation $x$, and transitional content $y$ (illustrated in Figure \ref{fig:causal graph}). In many cases, sentence order is independent of the specified content and directly establishes relations with transitional content. For example, we tend to use \textit{"firstly"} at the beginning and \textit{"finally"} at the end while composing a paragraph, regardless of what exactly is in between. 
This relation corresponds to path $c \to y$, and it should be preserved as an writing experience or habit. Meanwhile, there is a lot of transitional content that portrays the relations between referred papers based on the actual content, at this time, models need to analyze and use these relations. The corresponding path is $x \to y$. 

Though ideally, sentence order and document relation can instruct the generation of transitional content 
based on practical writing needs,
quite often, deep learning models are unable to trade off the influence of these two aspects correctly but prioritize sentence order. 
This can be attributed to the fact that sentence order information is easily accessible and learnable. 
In Figure \ref{fig:causal graph}, such relation corresponds to $c \to x \to y$. In this case, sentence order $c$ is the confounder that raises a spurious correlation with transitional content $y$. 
Although performing well on the training set, 
once a data distribution shift exists between the test set and training set where the test set focuses more on document relations,
the transitional content instructed by sentence order can be quite unreliable. 
In order to mitigate the impact of the spurious correlation, we need to cut off the path $c \to x$, enabling the model to generate transitional content based on the correct and reliable causality of both $c \to y$ and $x \to y$.

\subsection{Causal Intervention Module for Related Work Generation}
The proposed Causal Intervention Module for Related Work Generation (CaM) contains three parts.
\textbf{Primitive Intervention} performs causal intervention 
and preliminarily removes the spurious correlations between sentence order and transitional content.
\textbf{Context-aware Remapping} 
captures and fuses contextual information, facilitating the smoothing of the intervened embeddings.
\textbf{Optimal Intensity Learning} learns the best intensity of holistic causal intervention. 
The overall structure is demonstrated in Figure \ref{fig:overall model}.

\subsubsection{Primitive Intervention}  \label{primitive itv}
Based on the causal graph G shown in Figure \ref{fig:causal graph}, we first perform the following derivation using \textit{do}-calculus and \textit{backdoor criterion}.
\begin{equation}
\begin{aligned}
    p(y|do(x)) &= \begin{matrix} \sum_{\mathbf{c}} p(y|do(x),\mathbf{c})p(\mathbf{c}|do(x)) \end{matrix}\\
                &= \begin{matrix} \sum_{\mathbf{c}} p(y|x,\mathbf{c})p(\mathbf{c}|do(x))   \end{matrix}\\
                &= \begin{matrix} \sum_{\mathbf{c}} p(y|x,\mathbf{c})p(\mathbf{c})   \end{matrix} 
    \label{eq:do-opt}
\end{aligned}
\end{equation}
In short, the do-calculus is a mathematical representation of an intervention, and the backdoor criterion can help identify the causal effect of $x$ on $y$ \cite{pearl-causality}.
As a result, by taking into consideration the effect of each possible value of sentence order $c$ on transitional content $y$, $c$ stops affecting document relation $x$ when using $x$ to estimate $y$, which means path $c \to x$ is cut off (see the arrow-pointed graph in Figure \ref{fig:causal graph}). 
Next, we will explain how to estimate separately $p(y|x,\mathbf{c})$ and $p(\mathbf{c})$ using deep learning models and finally obtain $p(y|do(x))$.

Let
$E^{ori} = (e^{ori}_1, {e}^{ori}_2,..., {e}^{ori}_{\hat{M}})$ denote the input embeddings 
corresponding to $\hat{M}$-sized related work
and ${E}^{itv} = ({e}^{itv}_1, {e}^{itv}_2,..., {e}^{itv}_{\hat{M}})$ denote the output embeddings of Primitive Intervention.
We first integrate the sentence order information into the input embeddings:
\begin{equation}    \label{eq:order_concat}
    {e}^{odr(j)}_i =\mathrm{Linear} ({e}^{ori}_i \oplus {o}_j)
\end{equation}
${O} = \{{o}_j\}_{j=1}^s$ denotes the order information for each sentence, and $s$ is the total number of sentences in the generated related work and can be considered as a hyper-parameter. 
${e}^{odr(j)}_i$ denotes the order-enhanced embedding for the $i$-th word which corresponds to the $j$-th sentence in related work. We take ${o}_j = (\lg{(j+1)},\cdots,\lg{(j+1)})$ with the same dimension as ${e}^{ori}$. The linear layer (i.e., $\mathrm{Linear}$) further projects the concatenated embedding to ${e}^{odr}$ with the same dimension as ${e}^{ori}$. Accordingly, we have the estimation of $p(y|x,\mathbf{c}):={e}^{odr}$. Then, we use a feed-forward network and the output subsequence $E_{sub} = ({e}^{itv}_1, ..., {e}^{itv}_{i-1})$ to predict the sentence position probability of the current decoding word:
\begin{align}
    & {h}_i =\mathrm{Softmax}( \mathrm{FFN}(\mathrm{ReLU}(\begin{matrix} \sum^{i-1} E_{sub} \end{matrix})))
\end{align}
each $h_i^j\in{h}_i$ denotes the probability. Thus, we estimate the sentence position probability of each decoding word $p(\mathbf{c}):={h}$.

After obtaining the estimation of $p(y|x,\mathbf{c})$ and $p(\mathbf{c})$, the final embedding with primitive causal intervention can be achieved:
\begin{equation}
    \begin{matrix} {e}^{itv}_i = \sum^s_{j=1} {{e}^{odr(j)}_i \times h^j_i}, h^j_i \in {h}_i \end{matrix}
\end{equation}
where ${e}^{odr(j)}_i \times h^j_i$ multiplying sentence order probability with order-enhanced embeddings is exactly $p(y|x,\mathbf{c})p(\mathbf{c})$
in Equation \ref{eq:do-opt}. 
The summation for each position $j$ completes the last step of Primitive Intervention.
Since most transitions are rendered by start words, our approach CaM intervenes only with these words, that is part of ${e}^{itv} \in {E}^{itv}$ is equal to ${e}^{ori} \in {E}^{ori}$. For simplicity, we still use ${E}^{itv}$ in the following. 

\subsubsection{Context-aware Remapping}
Two problems may exist in Primitive Intervention: 1) The lack of trainable parts may lead to the mapping spaces of the intervened embeddings and the original ones being apart and obstructs the subsequent decoding process. 2) Intervention on individual words may damage the context along with the order-enhanced embedding. 
To solve these two problems, we propose the Context-aware Remapping mechanism.

First, we scan ${E}^{itv}$ with a context window of fixed size $n_w$:
\begin{equation}
\begin{aligned}
    {B}_i &= \mathrm{WIN}_{i:i+n_w}([{e}^{itv}_1, {e}^{itv}_2, ..., {e}^{itv}_{\hat{M}}])   \\
    &= ({e}^{itv}_i, ..., {e}^{itv}_{i+n_w}), i=1,...,\hat{M}-n_w
\end{aligned}
\end{equation}
where $\mathrm{WIN}(\cdot)$ returns a consecutive subsequence of ${E}^{itv}$ at length $n_w$. 
Then, we follow the process of {Multi-head Attention Mechanism} \cite{attentionallyouneed} to update the embeddings in ${B}_i$:
\begin{equation}
    \begin{aligned}
        {B}_i^{rmp} &= \mathrm{MultiHead}({B}_i,{B}_i,{B}_i) \\
        &=({e}^{rmp}_i, ..., {e}^{rmp}_{i+n_w})
    \end{aligned}
\end{equation}
Even though all embeddings in ${B}_i$ are updated, we only keep the renewed 
${e}^{rmp}_{i+(n_w/2)} \in {B}_i^{rmp}$ , and leave the rest unchanged. 
Since $\mathrm{WIN}(\cdot)$ scans the entire sequence step by step, every embedding will have the chance to update.
The output is denoted as 
${E}^{rmp} = ({e}^{rmp}_1, {e}^{rmp}_2, ..., {e}^{rmp}_{\hat{M}})$.

\subsubsection{Optimal Intensity Learning}  \label{optimal lr}
In many cases, there is no guarantee that causal intervention with maximum (unaltered) intensity will necessarily improve model performance, 
especially when combined with pre-trained models~\cite{nips-gpt3, lewis-etal-2020-bart}, as the intervention may conflict with the pre-training strategies. 
To guarantee performance improvement, we propose {Optimal Intensity Learning}.

By applying {Primitive Intervention} and {Context-aware Remapping}, we have three types of embeddings, 
${E}^{ori}$,${E}^{itv}$, and ${E}^{rmp}$. 
To figure out their respective importance to the final output, we derive the output intensity corresponding to each of them:
\begin{align}
    g^{ori}_i & = \sigma ({W}_{ori} \cdot {e}^{ori}_i)  \\
    g^{itv}_i & = \sigma ({W}_{itv} \cdot {e}^{ori}_i)  \\
    g^{rmp}_i & = \sigma ({W}_{rmp} \cdot {e}^{ori}_i)  \\
    c^{ori}_i, c^{itv}_i, c^{rmp}_i & = f_s([g^{ori}_i, g^{itv}_i, g^{rmp}_i])
\end{align}
$\sigma(\cdot)$ is the $\mathrm{sigmoid}$ function, 
$f_s(\cdot)$ is the $\mathrm{softmax}$ function.
Combining $c^{ori}_i, c^{itv}_i, c^{rmp}_i$, we can obtain the optimal intervention intensity and 
the final word embedding set ${E}^{opm} = ({e}^{opm}_1,..., {e}^{opm}_{\hat{M}})$ with causal intervention:
\begin{equation}
    {e}^{opm}_i = c^{ori}_i  {e}^{ori}_i + c^{itv}_i  {e}^{itv}_i + c^{rmp}_i  {e}^{rmp}_i
\end{equation}

\subsection{Fusing CaM with Transformer}
\label{fuse cam}
To derive an end-to-end causal generation model and ensure that the intervened information can be propagated, 
we choose to integrate CaM with Transformer~\cite{attentionallyouneed}.
However, unlike the RNN-based models that generate words recurrently \cite{rnn-text-summarization}, 
the attention mechanism computes the embeddings of all words in parallel, while the intervention is performed on the sentence start words.

To tackle this challenge, we perform vocabulary mapping on word embeddings before intervention and compare the result with sentence start token $\mathrm{[CLS]}$ 
to obtain ${Mask}$:
\begin{gather}
    I = \mathrm{argmax}[\mathrm{Linear}_{vocab}({E}^{ori})]  \\
    {Mask} = \delta(I, ID_{CLS}) 
\end{gather}
$I$ contains the vocabulary index of each word.
$\delta(\cdot)$ compares the values of the two parameters, and returns $1$ if the same, $0$ otherwise.
$Mask$ indicates whether the word is a sentence start word. 
Therefore, ${E}^{opm}$ can be calculated as:
\begin{equation}
    {E}^{opm} = {E}^{opm} \odot {Mask} + {E}^{ori} \odot (\sim{Mask})
\end{equation}
The $\odot$ operation multiplies each embedding with the corresponding $\{0,1\}$ values, and $\sim$ denotes the inverse operation. Note that we omit ${Mask}$ for conciseness in Section \ref{optimal lr}.
${Mask}$ helps restore the non-sentence-start word embeddings and preserve the intervened sentence-start ones.

As illustrated in Figure \ref{fig:overall model}, we put CaM between the Transformer layers in the decoder.
The analysis of the amount and location settings will be discussed in detail in Section \ref{sec:fuse stategy}. 
The model is trained to minimize the cross-entropy loss between the predicted $\hat{Y}$ and the ground-truth ${Y}$, $v$ is the vocabulary index for $w_i \in Y$:
\begin{equation}
    \mathcal{L} = - \begin{matrix} \sum_i^{\hat{M}} \log p_{i}^v({\hat{Y}}) \end{matrix}
\end{equation}

\begin{table}[!t] \fontsize{10}{12}\selectfont
    \centering
	\begin{tabular}{lll}
		\toprule  
		\textbf{Statistic}      & \textbf{S2ORC}        & \textbf{Delve}                 \\ \midrule
        Pairs \#                & 126k/5k/5k   & 72k/3k/3k \\
		source \#               & 5.02                  & 3.69                \\
		words/sent(doc) \#      & 1079/45               & 626/26                 \\
		words/sent(sum) \#      & 148/6.69              & 181/7.88           \\
		vocab size \#           & 377,431               & 190,381           \\ \bottomrule
	\end{tabular}
	\caption{Statistics of the datasets}  \label{table_datasets}
	\vspace{-3mm}
\end{table}

\begin{table*}[!t] \fontsize{10}{12}\selectfont
	\centering
	\begin{tabular}{ccccccc}
		\toprule
		\multirow{2}{*}{Model}              &\multicolumn{3}{c}{\textbf{S2ORC}} &\multicolumn{3}{c}{\textbf{Delve}}        \\  
		                                    & ROUGE-1   & ROUGE-2   & ROUGE-L   & ROUGE-1   & ROUGE-2   & ROUGE-L           \\ \hline
		\multicolumn{1}{c}{\textit{Extractive Methods}}     \\
		\multicolumn{1}{c}{TextRank}       & 22.36     & 2.65      & 19.73     & 25.25     & 3.04      & 22.14             \\
		\multicolumn{1}{c}{BertSumEXT}     & 24.62     & 3.62      & 21.88     & 28.43     & 3.98      & 24.71            \\
		\multicolumn{1}{c}{MGSum-ext}      & 24.10     & 3.19      & 20.87     & 27.85     & 3.95      & 24.28             \\ \hline
		\multicolumn{1}{c}{\textit{Abstractive Methods}}    \\
		\multicolumn{1}{c}{TransformerABS} & 21.65     & 3.64      & 20.43     & 26.89     & 3.92      & 23.64             \\
		\multicolumn{1}{c}{BertSumABS}     & 23.63     & 4.17      & 21.69     & 28.02     & 3.50      & 24.74             \\
		\multicolumn{1}{c}{MGSum-abs}      & 23.94     & 4.58      & 21.57     & 28.13     & 4.12      & 24.95             \\
		\multicolumn{1}{c}{GS}             & 23.92     & 4.51      & 22.05     & 28.27     & 4.36      & 25.08             \\
          \multicolumn{1}{c}{T5-base}            & 23.20     & 4.01      & 21.41     & 26.38     & 5.69      & 24.35             \\
          \multicolumn{1}{c}{BART-base}            & 23.36     & 4.13      & 21.08     & 26.96     & 5.33      & 24.42             \\
      \multicolumn{1}{c}{longformer}            & 26.00     & 4.96      & 23.20    & 28.05     & 5.20      & 25.65             \\
		\multicolumn{1}{c}{RRG}            & 25.46     & 4.93      & 22.97     & 29.10     & 4.94      & 26.29             \\
		\multicolumn{1}{c}{\textbf{CaM (ours)}}      & \textbf{26.65} & \textbf{5.40} & \textbf{24.62} & \textbf{29.31} & \textbf{6.17} & \textbf{26.61}  \\ \bottomrule
	\end{tabular}
	\caption{ROUGE scores comparison between our CaM and the baselines.}
	\label{table_overall}
	\vspace{-2mm}
\end{table*}

\section{Experiments}
\subsection{Datasets}
Following the settings in~\newcite{chen-etal-2021-capturing,target-aware-2022}, we adopt two publicly available datasets derived from the scholar corpora S2ORC \cite{lo-wang-2020-s2orc} and Delve \cite{aku-delve} respectively to evaluate our proposed method in related work generation. S2ORC consists of scientific papers from multiple domains, and Delve focuses on the computer domain. 
The datasets are summarized in Table \ref{table_datasets}, where the corresponding ratios of the training/validation/test pairs are detailed.



\subsection{Settings}
We implement our model with PyTorch on NVIDIA 3080Ti GPU. 
In our model, the maximum reference paper number is set to 5, i.e., $|D|=5$. 
We select the first $440/|D|$ words in each reference paper abstract and concatenate them to obtain the model input sequence. 
The total number of sentences in target related work is set to 6, i.e., $s=6$. 
We use beam search for decoding, with a beam size of 4 and a maximum decoding step of 200.
When fusing CaM with Transformer, the dimension of word embedding is set to 768, both attention heads number and layer number is set to 12, 
and the intermediate size is set to 3072. 
We use Stochastic Gradient Descent(SGD) as the optimizer with a learning rate 1e-2. 
To ensure desirable performance and save training costs, we utilized pretrained BERT~\cite{devlin-etal-2019-bert}.
We use ROUGE-1, ROUGE-2 and ROUGE-L on F1 as the evaluation metrics \cite{lin-2004-rouge}. 
Since we adopt exactly the same datasets (including the dataset settings) as RRG~\cite{chen-etal-2021-capturing} used, we directly use the results in the RRG paper for baseline comparison.

\subsection{Compared Methods}
We compare our CaM with the following eight state-of-the-art baselines, including both extractive and abstractive methods.
\subsubsection{Extractive Methods}
(1) \textbf{TextRank} \cite{mihalcea-tarau-2004-textrank}: A graph-based text ranking model that can be used in multi-document sentence extraction.
(2) \textbf{BertSumEXT} \cite{liu-lapata-2019-text-bertsumext}: An extractive document summarization model that extends BERT by inserting multiple [CLS] tokens.
(3) \textbf{MGSum-ext} \cite{jin-etal-2020-mgsumext}: A multi-granularity interaction network that jointly learns different semantic representations.
    
\subsubsection{Abstractive Methods}
(1) \textbf{TransformerABS} \cite{attentionallyouneed}: An abstractive summarization model based on Transformer with attention mechanism.
(2) \textbf{BertSumABS} \cite{liu-lapata-2019-text-bertsumext}: An abstractive model based on BERT with a designed two-stage fine-tuning approach.
(3) \textbf{MGSum-abs} \cite{jin-etal-2020-mgsumext}: A multi-granularity interaction network that can be utilized for abstractive document summarization.
(4) \textbf{GS} \cite{li-etal-2020-leveraging-graph-gs}: An abstractive summarization model that utilizes special graphs to encode documents to capture cross-document relations. 
(5) \textbf{T5-base} \cite{t5}: A text-to-text generative language model that leverages transfer learning techniques.
(6) \textbf{BART-base} \cite{lewis-etal-2020-bart}: A powerful sequence-to-sequence model that combines the benefits of autoregressive and denoising pretraining objectives.
(7) \textbf{Longformer} \cite{Beltagy2020Longformer}: A transformer-based model that can efficiently process long-range dependencies in text.
(8) \textbf{RGG} \cite{chen-etal-2021-capturing}: An encoder-decoder model specifically tailored for related work generation, which constructs and refines the relation graph of reference papers.

\subsection{Overall Performance}

It can be found in Table \ref{table_overall} that abstractive models have attracted more attention in recent years and usually outperform extractive ones. 
Among the generative models, pretrained model T5 and BART achieve promising results in our task without additional design. 
Meanwhile, Longformer, which is good at handling long text input, also achieves favorable results. 
However, the performance of these models is limited by the complexity of the academic content in the dataset.


Our proposed CaM achieves the best performance on both datasets. Due to fusing CaM with Transformer, 
its large scale ensures that our model can still effectively capture document relations without additional modeling.
Accordingly, CaM enables the model to obviate the impact of spurious correlations through causal intervention and promotes the model to learn more robust causalities to achieve the best performance.

\subsection{Ablation Study}
\begin{figure}[!htbp]
	\centering
	\includegraphics[width=0.85\linewidth]{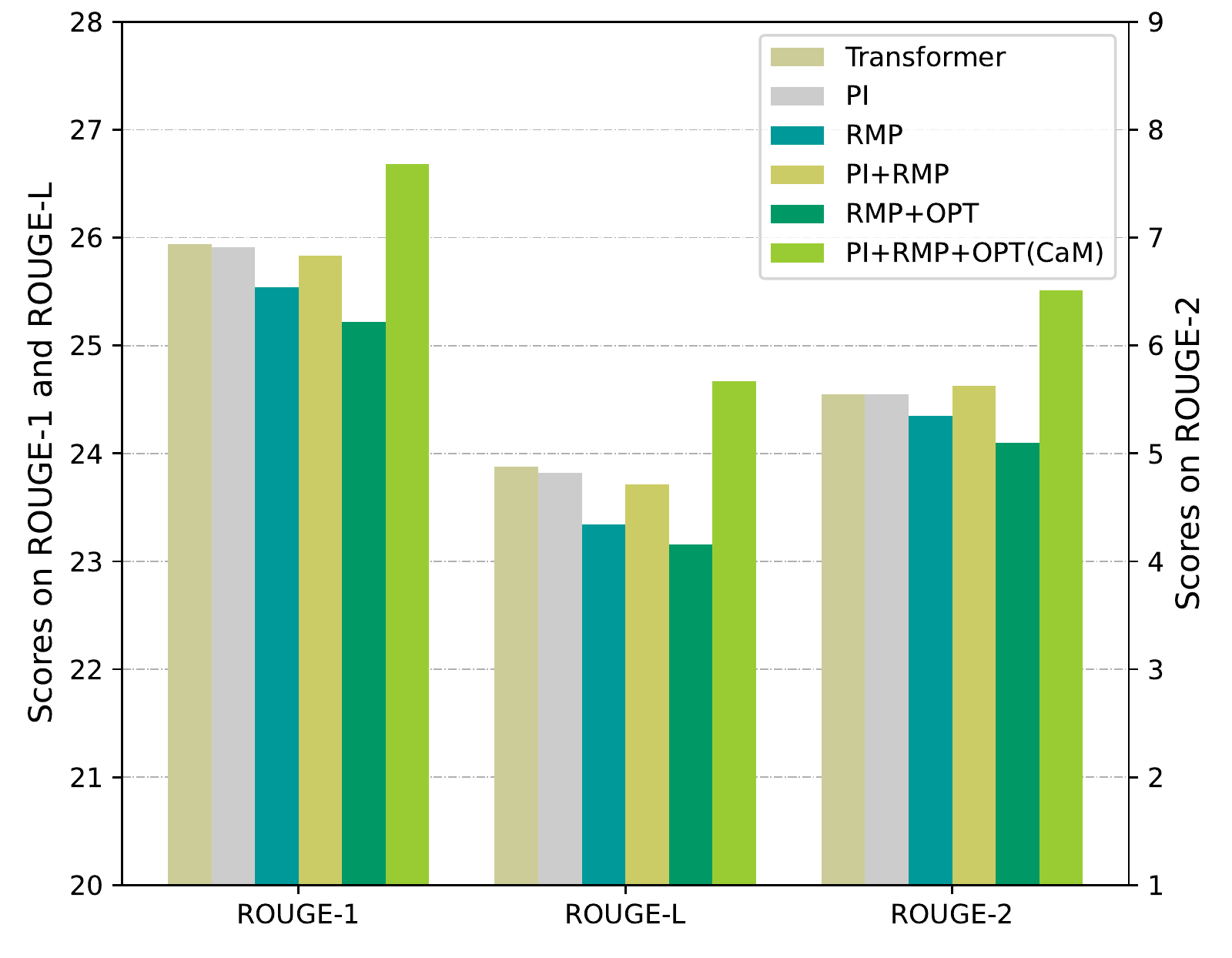}
	\vspace{-1mm}
	\caption{Ablation result on S2ORC.}
	\label{fig:abl_s2orc}
\end{figure}
\begin{figure}[!htbp]
	\centering
	\includegraphics[width=0.85\linewidth]{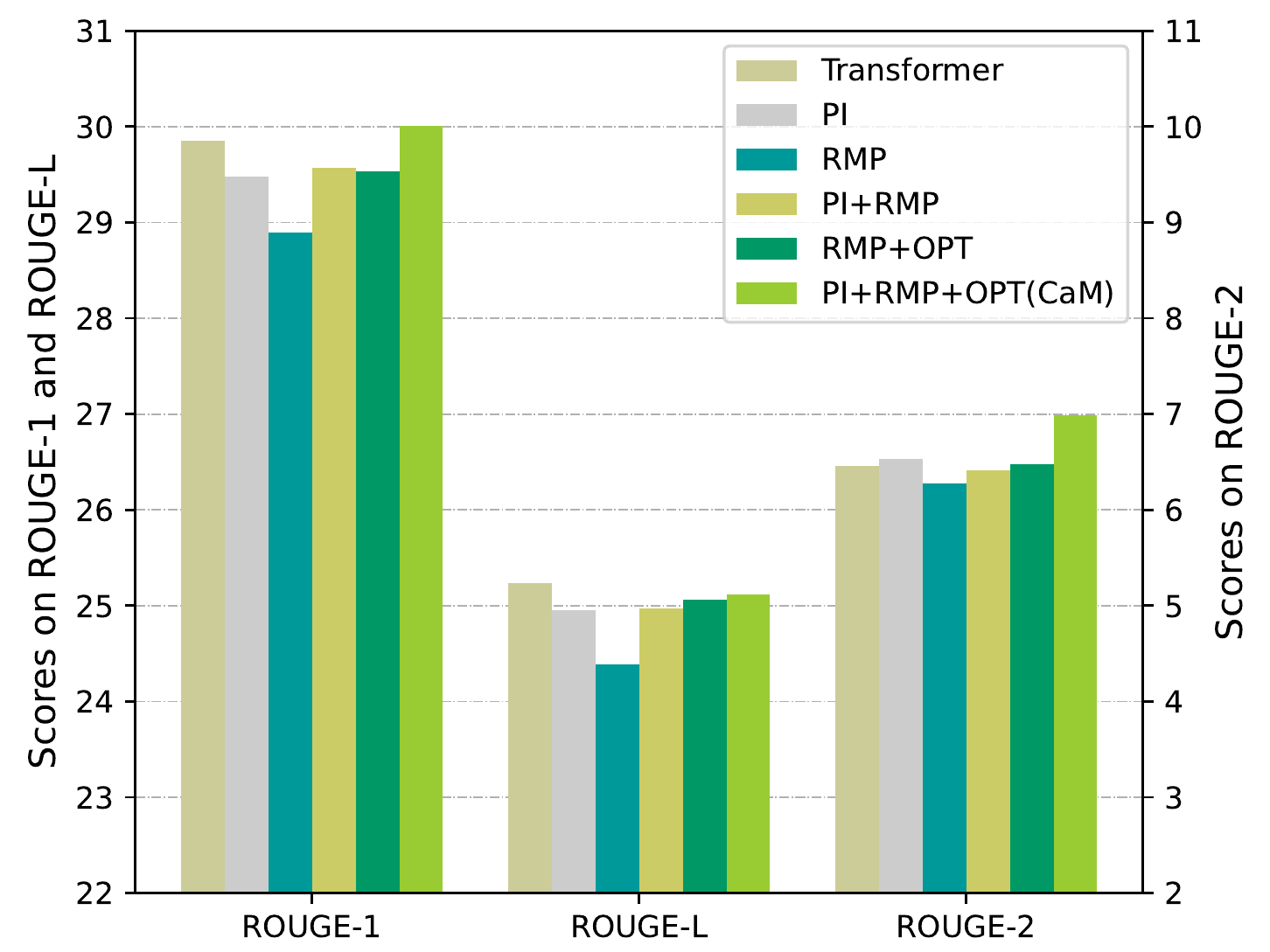}
	\vspace{-1mm}
	\caption{Ablation result on Delve.}
	\label{fig:abl_delve}
\end{figure}

To analyze the contribution of the different components of CaM, we separately control the use of Primitive Intervention (PI), Context-aware Remapping (RMP) and Optimal Intensity Learning (OPT). 
Figure \ref{fig:abl_s2orc} and Figure \ref{fig:abl_delve} show the performance comparison between different variants of CaM. 

First, it can be observed that the basic Transformer model already guarantees a desirable base performance. 
When only PI is used, the model generally shows a slight performance drop. 
PI+RMP outperforms RMP, showing the necessity of the PI and the effectiveness of RMP. 
PI+RMP+OPT achieves optimal results, indicating that OPT can effectively exploit the information across different representations.


\subsection{Human Evaluation}
\begin{table}[!t] \fontsize{10}{12}\selectfont
    \centering
	\begin{tabular}{llll}
		\toprule  
		\textbf{}      & \textbf{inf}        & \textbf{coh}   & \textbf{suc}                \\ \midrule
        \textbf{CaM}                & 2.21   & 2.38  & 2.01      \\
		\textbf{RRG}              & 2.07   & 2.10  & 2.05                \\
		\textbf{BERT}      & 2.11   & 1.97   & 1.92                 \\
		\toprule  
	\end{tabular}
	\caption{Human evaluation result}  \label{human_eval}
	\vspace{-3mm}
\end{table}
 We evaluate the quality of related works generated by the CaM, RRG, and BERT from three perspectives (informativeness, coherence, and succinctness) by randomly selecting forty samples from S2ORC and rating the generated results by 15 master and doctoral students on the three metrics (from 0 to 3, with higher scores indicating better results). As table \ref{human_eval} shows, our method achieves the best in informativeness and coherence, and the causal intervention makes coherence the most superior. However, succinctness is slightly lower than RRG, probably due to the output length limit. We will complete the human evaluation by more participants using the MTurk platform and report the evaluation results in the final version.

\subsection{Fusing Strategy Comparison} \label{sec:fuse stategy}
\begin{figure}[t]
	\centering
	\includegraphics[width=0.85\linewidth]{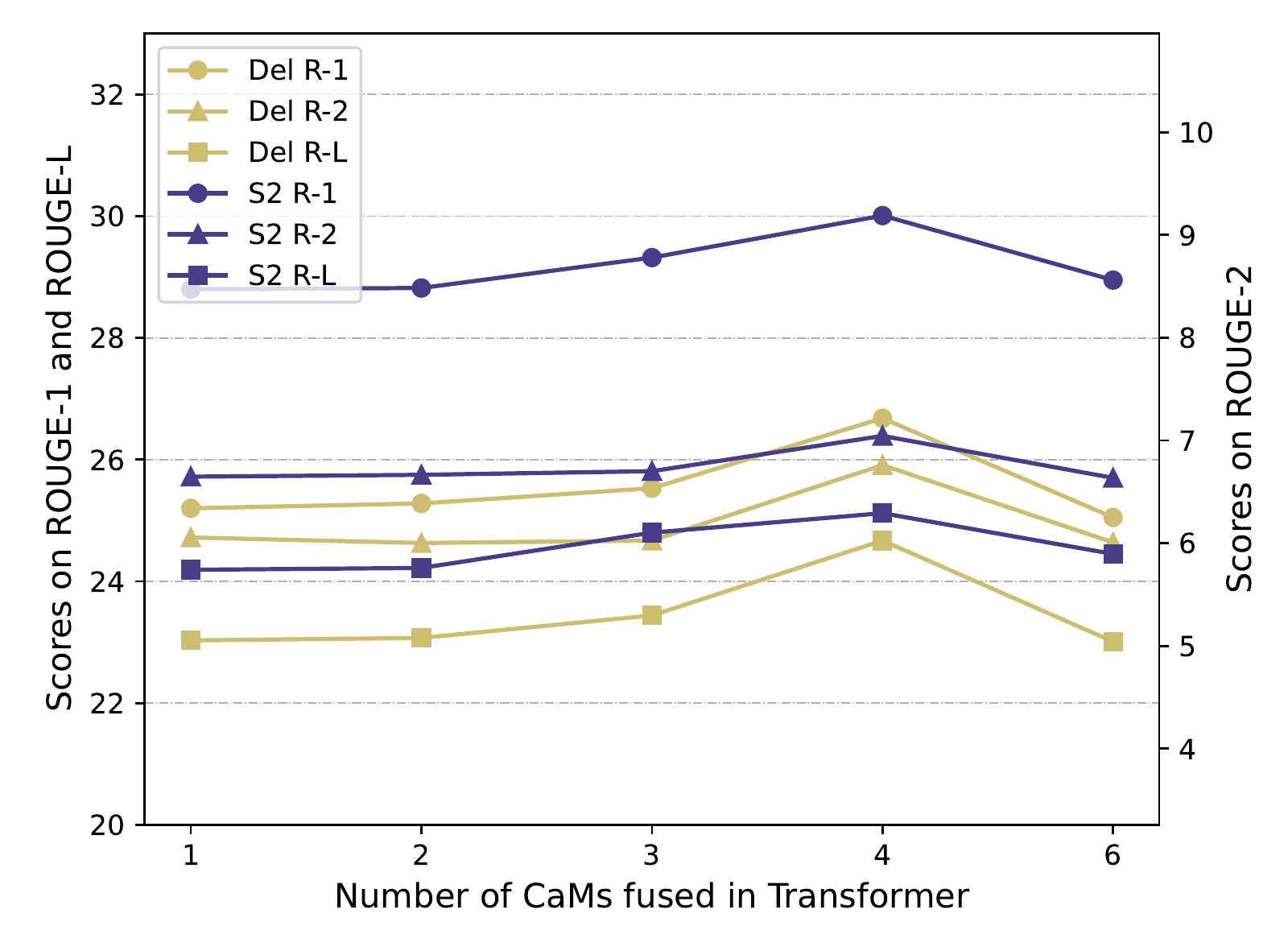}
	\vspace{-1mm}
	\caption{Performance analysis on the number of CaMs fused with Transformer.}
	\label{fig:param}
	\vspace{-2mm}
\end{figure}

In our setting, the base Transformer model consists of $12$ layers, so there are multiple locations to fuse a different number of CaMs. 
For each scenario, CaMs are placed evenly among the Transformer layers, and one will always be placed at the end of the entire model. 
The results of all cases are shown in Figure \ref{fig:param}.
It can be observed that the model performs best when the number of CaM is 4 both on S2ORC and Delve. 
With a small number of CaMs, the model may underperform the benchmark model and fail to achieve optimal performance due to the lack of sufficient continuous intervention. 
If there are too many CaMs, the distance between different CaMs will be too short, leaving an insufficient learning process for the Transformer layers, 
and this might cause the CaMs to bring the noise.

\subsection{Robustness Analysis}
To verify the robustness of knowledge learned by causal intervention, we designed two experiments on CaM and the base Transformer (TF).

\subsubsection{Testing with Reordered Samples}
We randomly select 50 samples (15 from S2ORC and 35 from Delve) and manually rearrange the order of the cited papers in each of them, 
as well as the order of their corresponding sentences.
Transitional content in related works is also removed since the reordering damages the original logical relations.

\begin{figure}[!htbp]
	\centering
	\includegraphics[width=1\linewidth]{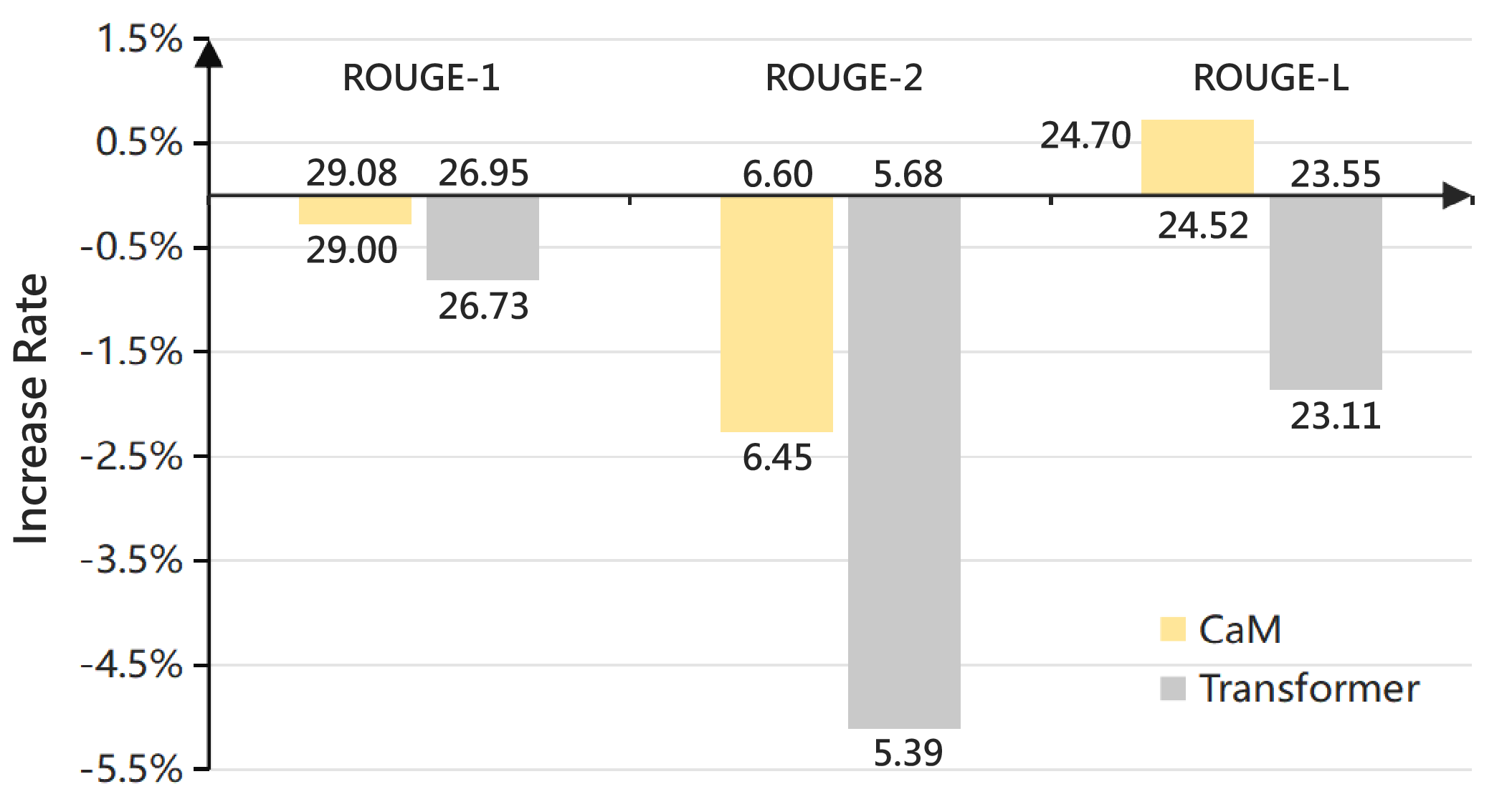}
	\vspace{-1mm}
	\caption{Comparison between Transformer and CaM on original and reordered samples.}
	\label{fig:robust0}
	\vspace{-1mm}
\end{figure}

It can be observed from Figure \ref{fig:robust0} that CaM has better performance regardless of whether the samples have been reordered or not. 
By switching to the reordered samples, the performance of Transformer decreases on all three metrics, 
but CaM only decreases on ROUGE-1 and ROUGE-2 at a much lower rate.
Particularly, compared to the Transformer, CaM makes improvement on ROUGE-L when tested with reordered samples.
The result indicates that CaM is able to tackle the noise disturbance caused by reordering, and the generated content maintains better coherence.

\subsubsection{Testing with Migrated Test Set}

\begin{figure}[!htbp]
	\centering
	\includegraphics[width=0.85\linewidth]{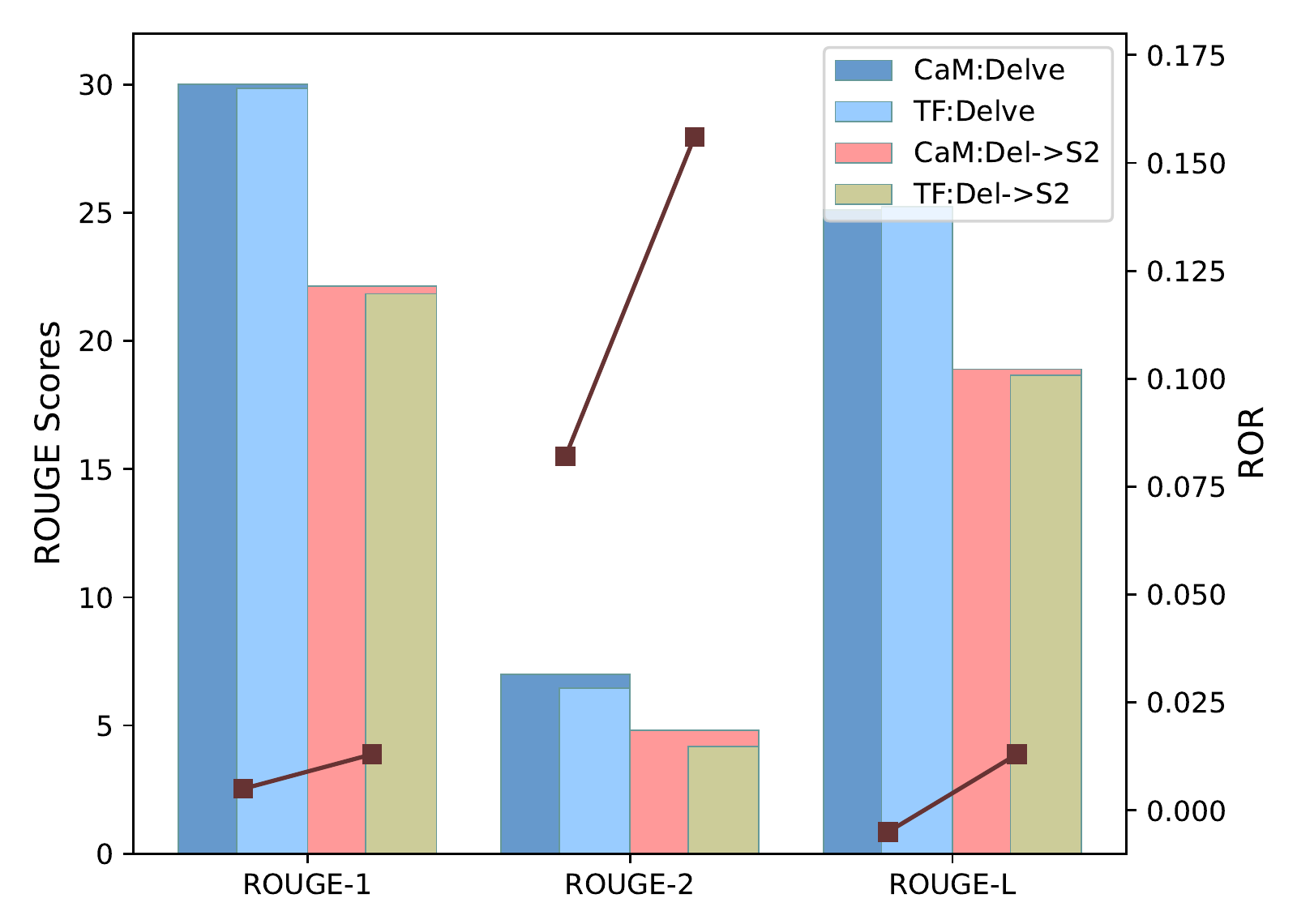}
	\caption{The result of migrating test set from Delve to S2ORC (trained on Delve).}
	\label{fig:robust1}
\end{figure}

We train the models on Delve and test them on S2ORC, which is a challenging task and significant for robustness analysis. 
As expected, the performances of all models drop, but we can still obtain credible conclusions. Since CaM outperforms Transformer initially, simply comparing the ROUGE scores after migrating the test set is not informative. 
To this end, we use \textit{Relative Outperformance Rate} (ROR) for evaluation:
\begin{equation}
    \mathrm{ROR} = (\mathrm{S_{CaM}} - \mathrm{S_{TF}}) / \mathrm{S_{TF}}
\end{equation}
$\mathrm{S_{CaM}}$ and $\mathrm{S_{TF}}$ are the ROUGE scores of CaM and Transformer, respectively. 
ROR computes the advantage of CaM over Transformer.

Figure \ref{fig:robust1} reports that CaM outperforms Transformer regardless of migrating from Delve to S2ORC for testing. In addition, comparing the change of ROR, we observe that although migration brings performance drop, CaM not only maintains its advantage over Transformer but also enlarges it. The above two experiments demonstrate that the CaM effectively learns causalities to improve model robustness.

\subsection{Causality Visualization} \label{sec:cv}
\begin{figure}[!htbp]
	\centering
	\includegraphics[width=1\linewidth]{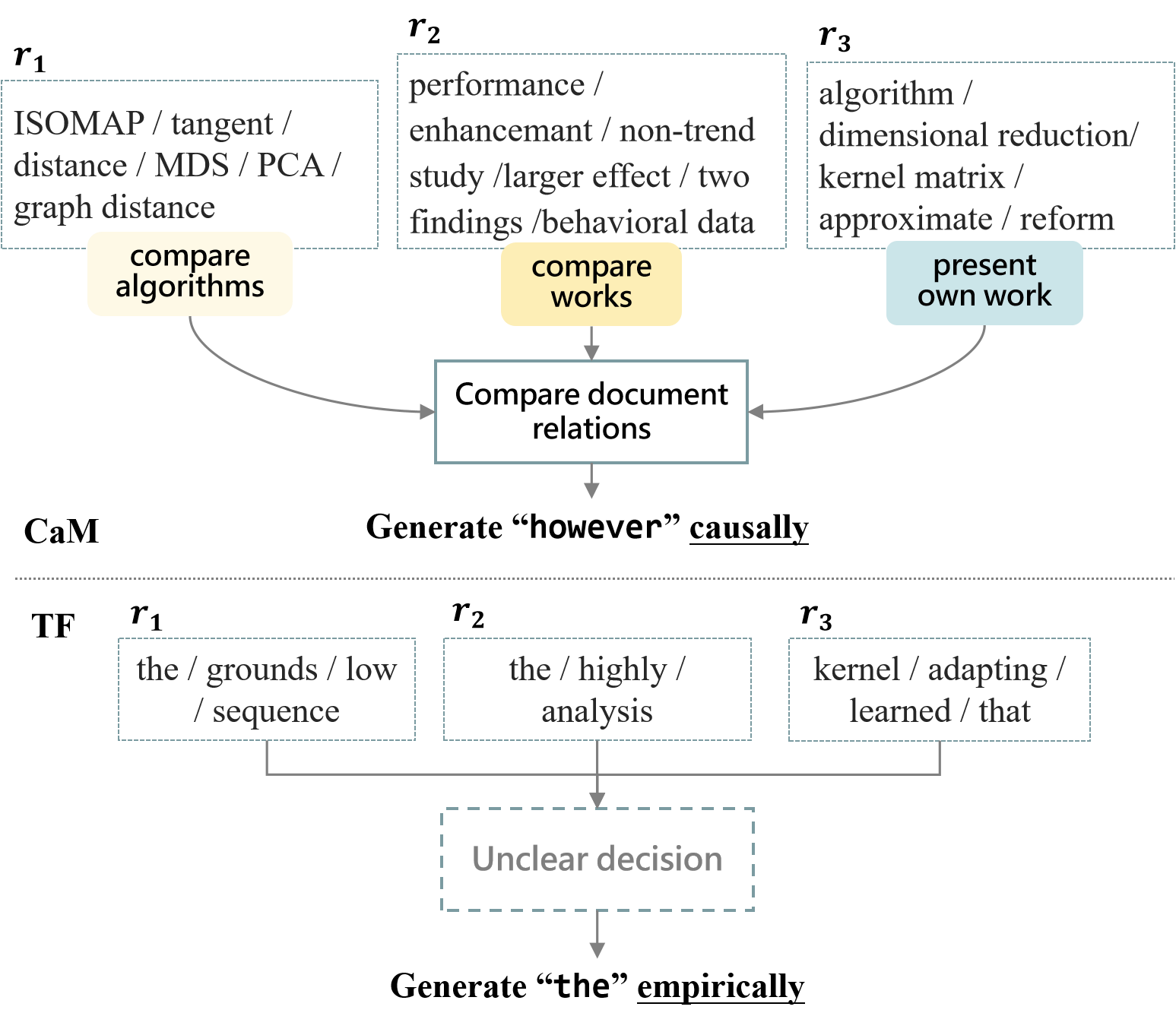}
	\vspace{-1mm}
	\caption{Visualization of the generating process within CaM and Transformer(TF).}
	\vspace{-2mm}
	\label{fig:ca_vi}
\end{figure}

To visualize how causal intervention worked in the generation process, we compare the related works generated by the base Transformer and CaM with a case study (full results in Table \ref{table:rw compare}). 
Specifically, we map their cross attention corresponding to {"however"} and {"the"} to the input content using different color shades 
(see Figure \ref{fig:ca_vis_com}) to explore what information of these two words rely on.
More details of the above two experiments can be found in Appendix \ref{vi analysis}.

We picked out the words that {"however"} and {"the"} focused on the most and 
analyzed the implications of these words in the context of the input. The results are shown in Figure \ref{fig:ca_vi}.
It can be found that the words highlighted by CaM have their respective effects in the cited papers.
When generating {"however"}, the model aggregates this information, comparing the relations between the documents and producing the correct result. 
However, there is no obvious connection between the words focused on by Transformer, hence there is no clear decision process after combining the information, and the generated word {"the"} is simply a result obtained from learned experience and preference. Through causality visualization, it can be observed very concretely how CaM improves model performance by conducting causal intervention.

\section{Conclusions}
In this paper, we propose a Causal Intervention Module for Related Work Generation (CaM) to capture causalities in related work generation. 
We first model the relations in related work generation using a causal graph.
The proposed CaM implements causal intervention and 
enables the model to 
capture causality.
We subtly fuse CaM with Transformer to obtain an end-to-end model to integrate the intervened information throughout the generation process.
Extensive experiments show the superiority of CaM over the latest models and demonstrate our method's effectiveness.

\section*{Limitations}
Although extensive experiments have demonstrated that CaM can effectively improve the performance of the base model, 
as mentioned above, since the intervention occurs on the sentence start words, 
it is inconclusive that CaM can bring improvement if the generation of sentence start words is inaccurate. 
That is, CaM can improve the effect of large-scale models or pre-trained models very well, 
but if it is to be combined with small-scale models and trained from scratch, 
then the effectiveness of the model might not be ensured. 
This will also be a direction of improvement for our future work.

\bibliography{acl2023}
\bibliographystyle{acl_natbib}

\appendix
\section{Related Work}
\subsection{Related Work Generation}
The related work generation task can be viewed as a variant of the multi-document summarization task, 
and its methods can be categorized as extractive or abstractive. 
Most of the early studies use extractive methods.
The work of \newcite{hoang-kan-2010-towards-automated-firstwork} is one of the first attempts. 
They propose a heuristic approach to generate general and specific content separately given a topic tree.
\newcite{wang-cited-text-span} train the model to extract cited text spans through a specific training set and use a greedy algorithm 
to select the most suitable candidate sentences to compose related works. 
Most recent studies focus on abstractive approaches. 
\newcite{xing-etal-2020-automatic-citation-context} use the citation context and the abstract of the cited papers together as inputs to generate citation text. 
\newcite{chen-etal-2021-capturing} construct a relation graph of the cited papers during the encoding process and update them iteratively. The relation graph is used as an auxiliary information for decoding.
The most recent work is done by \newcite{target-aware-2022}, 
in which they take the paper that related work belongs to as the target and employ a target-centered attention mechanism to generate informative related works. 
Besides the information from reference papers, they also use words that can represent the target paper as the input, so we did not compare this method as a baseline.

\subsection{Causal Intervention}
In recent years, causality theory has attracted increasing attention in various domains.
In the field of recommendation system, \newcite{multi-scen-causalint} use the causal graph to model multi-scenario recommendation and solve the problem of existing systems that may introduce unnecessary information from other scenarios.
\newcite{unbiased-recom-latent-confounder}
propose a framework for sequential recommendation that can perceive data biases by reweighing training data and using inverse propensity scores\cite{ips-score}.
In the field of natural language processing, \newcite{counterfactual-reasoning} introduce counterfactual reasoning into the sentiment analysis task and leverage the knowledge of both factual and counterfactual samples. 
\newcite{wang-culotta-2020-identifying} propose a method for identifying spurious correlations in the text classification task. The method extracts the words with the highest relevance to the category and uses an estimator to determine whether the correlation is a spurious correlation.

\section{Experiment Result for Causal Visualization}     \label{vi analysis}
In this section, we will give an extra analysis of the experiments introduced in Section \ref{sec:cv}.

\renewcommand\tabularxcolumn[1]{m{#1}}
\begin{table*}[ht!]\small
	\centering
		\begin{tabularx}{\textwidth}{c|X}
			\toprule
			Ground Truth     & Many dimension reduction techniques are proposed based on the vector forms, which are generally divided into two parts, linear and nonlinear. 
                                The classical methods of principal component analysis and multi-dimensional scaling are linear, 
                                since the outputs returned by these methods are related to the input patterns by a simple linear transformation. 
                                
                                \textcolor{olive}{\underline{\textbf{However}, when the input patterns lie on or near a low dimensional sub of the input space, }}
                                that is the structure of the data set may be highly nonlinear, then linear methods are bound to fail.
                                As the research for manifold learning, several graph-based nonlinear methods have been proposed, such as locally linear em. \\   \hline
			Transformer         & Reduction methods have been proposed on the dimensional space, such are divided into two categories: linear and nonlinear.
                                The first method are the component analysis, the dimensional of the methods are linear to the kernel data. 
                                The data of the input dimensional space are not linear to the large dimensional space. 
                                \textcolor{olive}{\underline{\textbf{The} data space dimensional of the data be the nonlinear, and  are not used. }}
                                The graph-based non-linear methods have been proposed. Including as the linear kernel, and entropy.           \\   \hline
			CaM        & Reduction methods have been proposed on the kernel space, such are divided into two categories: linear and nonlinear. 
                                The first approach component analysis are linear and dimensional analysis are based the kernel of the methods. 
                                Data of the input are not represented to the low dimensional space. 
                                
                                \textcolor{olive}{\ul{\textbf{However}, the data are not on a low dimensional space. The data space is more nonlinear, and the methods can not be used.}}
                                The graph-based nonlinear methods have been proposed. Including as the linear entropy.           \\ 
        \bottomrule
		\end{tabularx}
            \caption{Related works generated by CaM and Transformer. Analysis of the \textbf{bolded} words is in Section \ref{sec:cv}.}
		\label{table:rw compare}
\end{table*}

\subsection{Generated Related Work Comparison} 
From Table \ref{table:rw compare}, we can notice that CaM generates enriched content and its meaning is closer to ground truth compared to Transformer.
Crucially, when pointing out the problems of previous approaches and presenting the new ones(sentence marked in green), 
CaM correctly generates {"however"} at the beginning of the sentence and the entire sentence has a more accurate expression, 
making the transitions more seamless.
But Transformer only generates a very high-frequency word {"the"} at the same position. 
It can be perceived that in this process Transformer is not making effective decisions, but simply generating with preference and experience.

\begin{figure*}[t]
    \centering
    \includegraphics[width=1\linewidth]{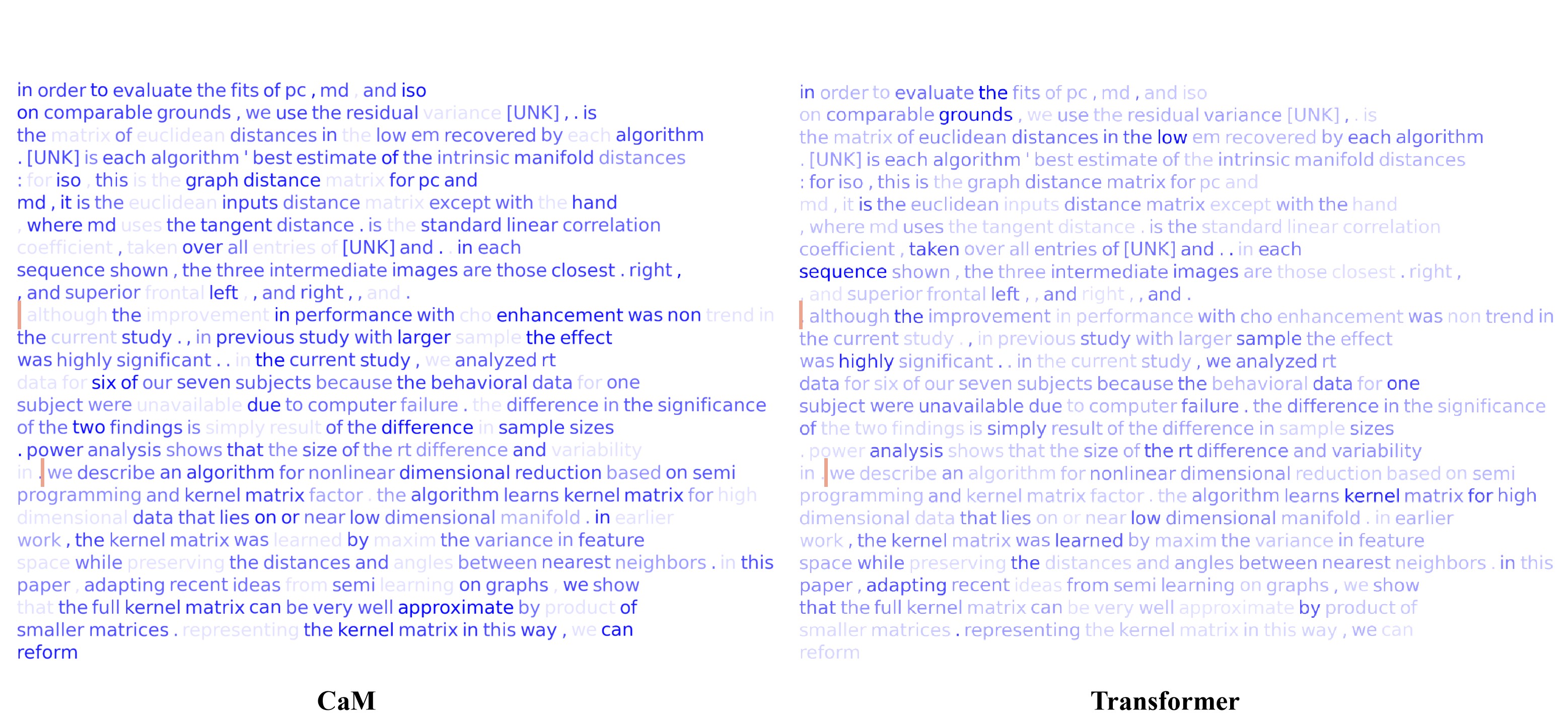}
    \caption{Raw visualization result from CaM on the word {"however"} and Transformer on the word {"the"}.}
    \label{fig:ca_vis_com}
\end{figure*}

\subsection{Visualization Result Analysis on Full Text}    
Figure \ref{fig:ca_vis_com} visualizes the cross attention of words {"however"} and {"the"} in CaM and Transformer.
Different cited papers are split with vertical lines.
The deeper blue color denotes the higher attention received by the input source word.
Judging from the overall coloring situation, we can find that in CaM, there is more deep blue text, as well as more light-colored text.
This means the information that {"however"} focuses on is more targeted and more important, and CaM is capable to produce correct content by accurately capturing document relations and avoid distractions from the confounder.
In the result of Transformer, both light and deep blue text become less visible, and the coverage of normal blue increases greatly, 
indicating that {"the"} focuses on a wider range of information but lacks emphasis. 
It indicates that the decision process in Transformer is unclear and ineffective.

Detailed analysis of the exact words they focus on and the decision process of the models is presented in Section \ref{sec:cv}.

\end{document}